\documentclass[lettersize,journal]{IEEEtran}
\usepackage{amsmath,amsfonts}
\usepackage{algorithmic}
\usepackage{algorithm}
\usepackage{array}
\usepackage[caption=false,font=normalsize,labelfont=sf,textfont=sf]{subfig}
\usepackage{textcomp}
\usepackage{stfloats}
\usepackage{url}
\usepackage{verbatim}
\usepackage{graphicx}
\usepackage{cite}
\usepackage{multirow}
\usepackage{array}  
\usepackage{makecell}

\begin{document}

\title{TransLandSeg: A Transfer Learning Approach for Landslide Semantic Segmentation Based on Vision Foundation Model}

\author{Changhong Hou, Junchuan Yu, Daqing Ge, Liu Yang, Laidian Xi, Yunxuan Pang, and Yi Wen

}



\maketitle

\begin{abstract}
Landslides are one of the most destructive natural disasters in the world, posing a serious threat to human life and safety. The development of foundation models has provided a new research paradigm for large-scale landslide detection. The Segment Anything Model (SAM) has garnered widespread attention in the field of image segmentation. However, our experiment found that SAM performed poorly in the task of landslide segmentation. We propose TransLandSeg, which is a transfer learning approach for landslide semantic segmentation based on a vision foundation model (VFM). TransLandSeg outperforms traditional semantic segmentation models on both the Landslide4Sense dataset and the Bijie landslide dataset. Our proposed adaptive transfer learning (ATL) architecture enables the powerful segmentation capability of SAM to be transferred to landslide detection by training only 1.3\% of the number of the parameters of SAM, which greatly improves the training efficiency of the model. Finally we also conducted ablation experiments on models with different ATL structures, concluded that the deployment location and residual connection of ATL play an important role in TransLandSeg accuracy improvement.
\end{abstract}


\section{Introduction}
\IEEEPARstart{O}{nce} a landslide occurs, it often brings catastrophic destruction to humanity, demolishing houses, roads, farmland, and even endangering human lives [1]. From 1995 to 2014, there were a total of 3,876 landslide events worldwide, resulting in 163,658 deaths and 11,689 injuries [2]. Therefore, timely and accurate identification of landslides is of significant importance in reducing loss of life and property [3]. The early stages of a landslide investigation primarily relied on field surveys conducted by investigators. Although this method is highly accurate, it requires a significant amount of manpower and resources, has a long working cycle, and is not suitable for wide-area landslide investigations [1]. The fast growth of remote sensing technologies [4], such as high-resolution images [5], interferometric synthetic aperture radar (InSAR) [6], and others, has made it possible to collect a lot of data that can be used to find landslides. Investigators can utilize remote sensing data and employ visual interpretation methods to extract and identify landslide information [3], [7]. However, this approach heavily relies on expert knowledge and experience [3], [8], [9]. The detection efficiency still needs further improvement. Researchers are focusing on machine learning methods to enhance the efficiency of landslide detection [10], [11], [12]. These methods can be primarily classified into two categories. The first is pixel-based method, where landslide detection is achieved by classifying each pixel of a high-resolution image [13], [14]. Another approach is the object-based method, where landslide detection is achieved by classifying attribute features of remote sensing data, such as spectral, spatial, and texture [15], [16]. Although these machine learning methods have overcome the drawback of slow visual interpretation speed, they heavily rely on manually determining feature thresholds [1], [11].

In recent years, deep learning-based models for semantic segmentation and object detection have been increasingly applied in the field of landslide detection and have become mainstream technology. Firstly, the emergence of convolutional neural network (CNN) led to the development of a series of semantic segmentation models such as FCN [17], UNet [18], DeepLabv3+ [19], and object detection models such as Mask R-CNN [20] and Yolo [21]. These models have been applied in the field of landslide detection [22], [23], significantly enhancing the efficiency and accuracy of large-scale landslide detection [24], [25], [26]. Besides remote sensing imagery, digital elevation models (DEMs) and InSAR data can also provide useful information for landslide detection. A lot of CNN-based models for multi-source data fusion have been proposed [27], [28]. These methods have enhanced the accuracy of landslide detection. However, due to the fact that CNN expands the receptive field by stacking convolutions, they have certain limitations when it comes to extracting global information [29], [30]. The Transformer, based on the architecture of Multi-Head Self-Attention (MSA) [31], captures global contextual information during the initial feature extraction process and overcomes the limitations of CNN models [32]. The ViT-based model has been applied to landslide detection and has achieved superior performance compared to CNN models [29], [30], [33].

Research indicates that when the pre-training dataset is sufficiently large, the performance of the Transformer model will significantly surpass that of CNN models [31], [34]. However, training foundation models based on the Transformer architecture on large datasets requires a significant amount of computational resources. Additionally, the number of model parameters also grows exponentially. These are challenges for researchers [34]. Fortunately, since 2023, significant progress has been made in foundation models based on the Transformer architecture, such as BeRT [35], Pangu [36], and Chat-GPT. Their appearance solves the difficulties for researchers in training foundation model from scratch. The Segment Anything Model (SAM) has gained widespread attention as one of the VFMs in the field of image segmentation [37], [38], [39].SAM has been proven to possess successful segmentation abilities in various natural settings [37], [40], [41]. However, the training data for SAM consists of natural images and lacks remote sensing images. Due to the complex spectral and spatial characteristics of landslides, as well as their significant variations in scale and morphology [29], we believe that SAM is incapable of completing the landslide segmentation task. Therefore, our problem consists of transferring SAM's powerful segmentation capability to landslide tasks while maintaining low computational power and an efficient training process.

In this paper, we propose TransLandSeg, which is a transfer learning approach for landslide semantic segmentation based on a VFM. TransLandSeg employs adaptive transfer learning (ATL) to transfer the segmentation capability of SAM to downstream tasks of landslide segmentation. Inspired by the excellent parameter-efficient fine-tuning techniques in Natural Language Processing (NLP) [42], [43], we introduce a new set of trainable parameters into the original network to learn different features and inject the learned features into the original network to satisfy downstream tasks [44], [45].Our method achieves excellent performance on both the Landslide4Sense dataset and the Bijie landslide dataset. The contribution of this work can be summarized as follows:

\begin{enumerate} 
\item{we propose a new landslide segmentation method, TransLandSeg, which utilizes the ATL approach to migrate the powerful segmentation capability of SAM to solve the downstream landslide segmentation task.} 
\item{we conducted an extensive evaluation of the performance of TransLandSeg with other semantic segmentation models in different slide datasets. Our methodology still goes beyond existing methodologies and achieved excellent performance in these downstream missions.} 
\item{we explored and compared the impact of different ATL structures on the accuracy of TransLandSeg and got the best structure of TransLandSeg.}
\end{enumerate}

The rest of the paper is organized as follows: Section II provides a comprehensive review of related work in this area. Section III explains the proposed TransLandSeg model in detail. Section IV gives quantitative and qualitative results, including ablation experiments. In Section V, we explore and compare the effects of different ATL structures on the accuracy of TransLandSeg. Section VI concludes the paper by summarizing the main conclusions of the paper.

\section{Related Works}
\subsection{Landslide Detection Model}
Deep learning-based semantic segmentation and target detection models have become mainstream techniques in the field of landslide detection. Firstly, semantic segmentation models based on CNN are widely used for landslide detection. Among them, Ghorbanzadeh et al. [46] used CNN to recognize landslides in the southern part of Rasuwa district in Nepal and compared it with ANN, SVM, and RF, and the results showed that the deep learning method was significantly better than machine learning. The UNet network using residual connections has a larger receptive field and extracts global features more abstractly and essentially. With its excellent segmentation effect, it is widely used in landslide detection. Soares et al. [18] used the UNet model to automatically segment landslides in the city of Nova Friburgo in the mountainous area of Rio de Janeiro, Brazil, and achieved better results. He et al. [47] improved the DeepLabv3+ model with a spatial pyramid pool and codec structure. The results show that the improved DeepLabv3+ model has better extraction results and significantly improves the accuracy of landslide extraction compared with advanced methods such as UNet and PSPNet. Wang et al. [8] proposed the DPANet network and conducted experiments in a typical alpine valley area. Compared with the PSPNet model, the DPANet landslide detection accuracy improves by 4\% of OA and 18\% of PA and has good robustness to recognize complex landslides. The target detection model was similarly applied to landslide detection. Ju et al. [20] used Mask R-CNN to recognize old loess landslides on Google Earth imagery and compared it with RetinaNet and YOLOv3, respectively. The results show that Mask R-CNN is more suitable for recognizing old loess landslides. Li et al. [21] used YOLOv4 as the basic framework and the MobileNetv3 model as the backbone, and the modified YOLOv4 model improved the efficiency of landslide detection. Besides remote sensing images, multi-source data fusion models based on CNN have also been proposed. Ji et al. [24] used DEM data as an additional channel based on a CNN model that introduces an attention mechanism to improve the detection accuracy of landslides. Liu et al. [11] proposed FFS-Net, which fuses the terrain features extracted from DEM data with texture and shape features. Compared with UNet and DeepLabv3+, FFS-Net improves the accuracy of landslide detection. Liang et al. [27] used the UNet model to construct a landslide detection model based on InSAR deformation maps. The work of Zhou et al. [28] added a channel attention mechanism that models each interferogram separately. This makes InSAR timing analysis more accurate.

Unlike CNN, Transformer forgoes recursion and convolution, and it is a neural network model based on MSA mechanisms [32]. Its structure also dictates that it is more focused on capturing global contextual information than CNN [32]. In 2017, Transformer was first successfully applied in NLP [32] and then widely used in the field of computer vision (CV) [31]. Vision Transformer (ViT) and variants are beginning to be applied to landslide detection. Tang et al. [7] proposed SegFormer for landslide detection. SegFormer is based on the ViT structure and introduces overlapping patch embedding to capture the interactions between neighboring image patches. A simple MLP decoder and sequence reduction are also introduced to improve the efficiency of landslide detection. The ShapeFormer model was proposed by Lv et al. [29]. ShapeFormer is a semantic segmentation model based on the Pyramid Vision Transformer (PVT) structure. It lets you get more information about edges by extracting features from nearby elements at different sizes. The ShapeFormer model has better performance on the Bijie dataset and the Nepal dataset. Yang et al. [30] proposed a segmentation network with Transformer in ResU-Net and an embedded attention mechanism in the decoder for better fusion of Transformer and CNN feature mapping. Fu et al. [48] used Swin Transformer as a backbone for Mask R-CNN to detect Haiti earthquake landslide images through data augmentation and transfer learning.

Although the Transformer model requires more computational resources and storage space and is more difficult to train than the traditional CNN model, the hierarchical feature extraction strategy based on a convolutional kernel has some limitations in modeling the global information of an image. The Transformer model, based on the Mechanism of Self-Attention (MSA), provides a distinct perspective for global information modeling and thus shows great potential for implementation in the field of landslide detection [29], [30].

\subsection{Foundation Models}
Transformer's superior scalability makes it possible to build large-scale models with billions of parameters. Since 2023, foundation models represented by Chat-GPT have emerged, marking a new stage in the development of artificial intelligence (AI). Foundation models first made a breakthrough in NLP. They achieved amazing performance compared to previous methods. Foundation models have likewise emerged in the field of CV, including SAM [37] and so on.

 SAM is a VFM developed by Meta AI. It was pre-trained on the largest segmentation dataset (SA-1B), with 11 million images and more than 1 billion masks [37]. Large training data enables SAM to extract rich semantic features, and it can transfer zero-shots to new image distributions and tasks, which ensures its generalization potential in a variety of downstream scenarios. Peng et al. [49] proposed fine-tuning SAM efficiently by parameter space reconstruction (SAM-PARSER) to adapt to three different downstream scenarios. Wu et al. [50] proposed Medical SAM Adapter (MSA), which demonstrated superior performance on the 19 medical image segmentation tasks showing superior performance.

\subsection{Parameter-efficient fine-tuning}
Currently, there are two main fine-tuning paradigms for VFM [49]. One is full fine-tuning, i.e., tuning the entire parameters of the model, which requires a lot of computational resources and costs, but the gain is very limited [51]. And as the size of the model grows, this approach becomes less and less feasible [44]. Another approach is parameter-efficient fine-tuning(PEFT) [43]. i.e., the model is fine-tuned by adding or modifying a limited number of parameters of the model [45]. Drawing on the NLP field, PEFT are applied to the CV field. Bahng et al. [52] tuned a pre-trained model by modifying the original input pixel space. Jia et al. [53] proposed visual prompt tuning (VPT) to adjust the converter model for downstream visual tasks.

Among the PEFT, due to its plug-and-play, small parameters, and significant effect, adapter-tuning is widely used to fine-tune VFM [44]. The main idea of adapter-tuning is to add an adaptive module of size d to the original VFM of size  $\mathcal{M}$, $ d\ll \mathcal{M}$. Satisfactory results in terms of both efficacy and efficiency can be achieved without the need to fine-tune the full model [44]. Research shows that adapter-tuning is more effective than complete fine-tuning because it avoids catastrophic forgetting and can better generalize to out-of-domain scenarios with a small dataset. Recent studies have shown [39],[51],[54] that adapter-tuning can be better adapted to various downstream computer vision tasks. Therefore, we believe that adapter-tuning is the most suitable technique to bring SAM into the field of landslide detection.  

\section{Methods}
We have developed the TransLandSeg model to address the downstream task of landslide segmentation. In the following sections, we will first present an overview of the model's overall structure, followed by a detailed explanation of adaptive transfer learning (ATL).
\subsection{TransLandSeg model}
The TransLandSeg model, as shown in Figure 1(a), consists of three main modules: image encoder, mask decoder, and adaptive transfer learning (ATL) layer.

{\bf Image Encoder}: We have retained the original Image Encoder from SAM, which is a ViT pre-trained using Masked Auto Encoders (MAE). The specific model is a ViTH/16 model, which consists of a 14×14 window attention and 4 equidistant global attention blocks. In this paper, we select the large version of the SAM architecture, which consists of a backbone composed of 24 layers of Transformer blocks. The image is downsampled using an image encoder and then input into a mask decoder.The computation process of the image encoder is as follows:

\begin{align} 
\label{deqn_ex1}
x' &= \varepsilon \left( x \right) \\
\Phi_1 &= \tau_1\left( x' \right) \\
\Phi_i &= \tau_i\left( \Phi_{i-1} \right) \left ( 2\leq i\leq 24\right )
\end{align}
Where $x$ is the input image, $\varepsilon$ is patch embedding, $x'$ is the output of $x$ after passing through the patch embedding layer, $\Phi_i$ is the output of the i-th layer of the Transformer block, and $\tau_i$ is the i-th layer of the Transformer block.

\textbf{Mask Decoder}: We have retained the original mask decoder from SAM, which consists of a Transformer. The features extracted by the image encoder are first decoded using the Transformer, and then through the upsampling module $\mu$ and the MLP module $\rho$ to obtain the final mask results. The process is described as follows:
\begin{equation} 
\label{deqn_ex2} \psi \left ( \Phi \right )=\rho \left ( \mu \left ( {\tau }_{D}\left ( \Phi \right )\right )\right )
\end{equation}
Where $\psi$ is the output of the mask decoder, $\Phi$ is the output of the image encoder, and ${\tau }_{D}$ is the Transformer block in the mask decoder.

\textbf{ATL Layer}: The ATL method is based on the adapter-tuning parameter-efficient fine-tuning strategy. To be precise, an ATL layer is inserted between each Transformer block of the image encoder to learn specific knowledge. The output $\Phi_{i-1}$  of the previous Transformer block is fused with the output $\varTheta \left ( {\Phi }_{i-1}\right )$ of the ATL layer using residual connections and put into the next Transformer block. The training process is shown in Fig. 1(a), where the pre-trained weights of the ViT-L version of SAM are loaded first. Only the ATL layer and mask decoder are trained, and the Transformer layer and patch embedding are frozen. The detailed process is as follows:

\begin{align} 
\label{deqn_ex3}
\varOmega  \left ( X\right )&=\varTheta \left ( X\right )+X\\
{\Phi }_{1}&={\tau }_{1}\left ( {\varOmega  }_{tune}\left ( x'\right )\right )\\
{\Phi }_{i}&={\tau }_{i}\left ( {\varOmega  }_{tune}\left ( {\Phi }_{i-1}\right )\right )     \left ( 2\leq i\leq 24\right )\\
M&= {\psi }_{tune}\left ( {\Phi}_{24}\right )
\end{align}
Where $\varOmega$ denotes the output of the residual connection, $\varTheta$ denotes the ATL layer, the subscript tune indicates that the parameters of the module are trainable during the training process, and \textit{M} denotes the final output of the mask result.The training process is listed in Algorithm 1.
\begin{algorithm}
\caption{TransLandSeg Model Training}\label{alg:alg1}
\begin{algorithmic}[1]
\REQUIRE $N$: Image Encoder, Mask Decoder, ATL

\REPEAT
    \STATE Load SAM parameters to Image Encoder and Mask Decoder
    \STATE Freeze Image Encoder and train Mask Decoder and ATL
\UNTIL{Loss of $N$ is minimal}
\end{algorithmic}
\end{algorithm}

\begin{figure*}[!t] 
\centering 
\includegraphics[width=7.5in]{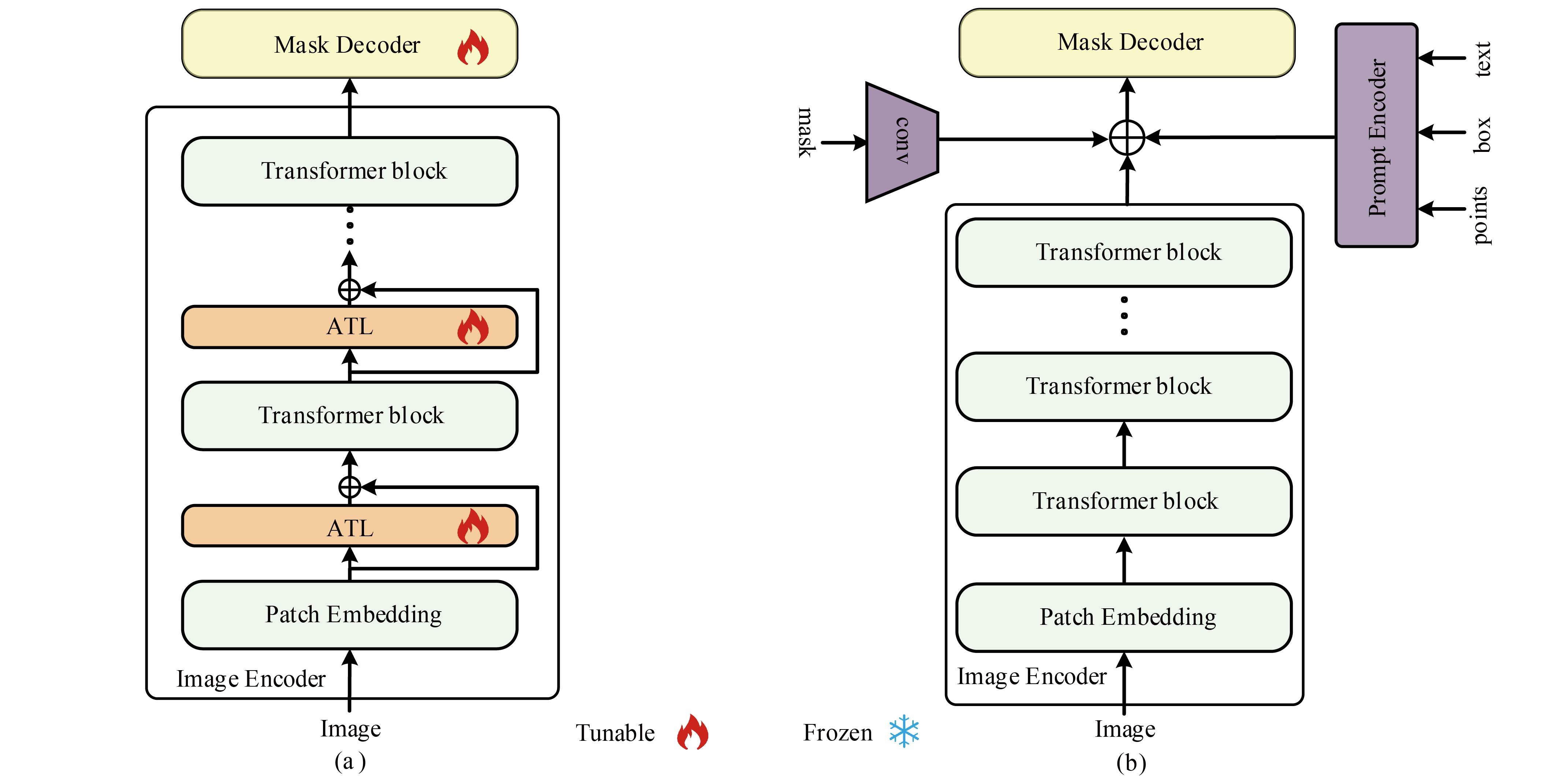} 
\caption{Structure of the proposed TransLandSeg and Segment Anything Model (SAM).} 
\label{fig1} 
\end{figure*}

\subsection{Adaptive Transfer Learning}
Our proposed ATL structure is shown in Fig. 2(a). It is composed of two fully connected layers and a nonlinear activation function. The ATL model employs a bottleneck structure to limit the number of parameters. It consists of a downward projection layer with parameters $ W_{down}\in {\mathbb{R}}^{\mathcal{M}\times d} $, a upward projection layer with parameters $W_{up}\in{\mathbb{R}}^{d\times\mathcal{M}}$, $d$ is the intermediate dimension and satisfies $d\ll \mathcal{M}$, $\mathcal{M}$ denotes number of output channels. In addition, MidLay is set up between the two projection layers to realize the nonlinear characteristics, and two kinds of MidLay units are set up in this paper. One is MidLay\_m, consisting of a fully connected layer and a GeLu activation function, as in Fig. 2(b);another is MidLay\_c, consists of a convolutional layer, a LayerNorm layer and a GeLu activation function layer,as in Fig. 2(c). The TransLandSeg model selects 1 MidLay\_m as the MidLay of the ATL. Assuming that the input features obtained by the ATL layer from the Transformer layer are ${\Phi }_{i}$, the output result after ATL computation is ${\varTheta }_{i}$, which is specifically expressed as:

\begin{equation} 
\label{deqn_ex4} {\varTheta }_{i}={\mathcal{L}}_{up}\left ( \mathcal{G}\left ( \mathcal{L}\left ( {\mathcal{L}}_{down}\left ( {\Phi }_{i}\right )\right )\right )\right ) 
\end{equation}
Where $\mathcal{L}_{down}$ is the downward projection, $\mathcal{L}_{up}$ is the upward projection, and $\mathcal{L}$ is the fully connected layer. $\mathcal{G}$ is the GELU activation function.

\begin{figure}[!t]
\centering
\includegraphics[width=2.5in]{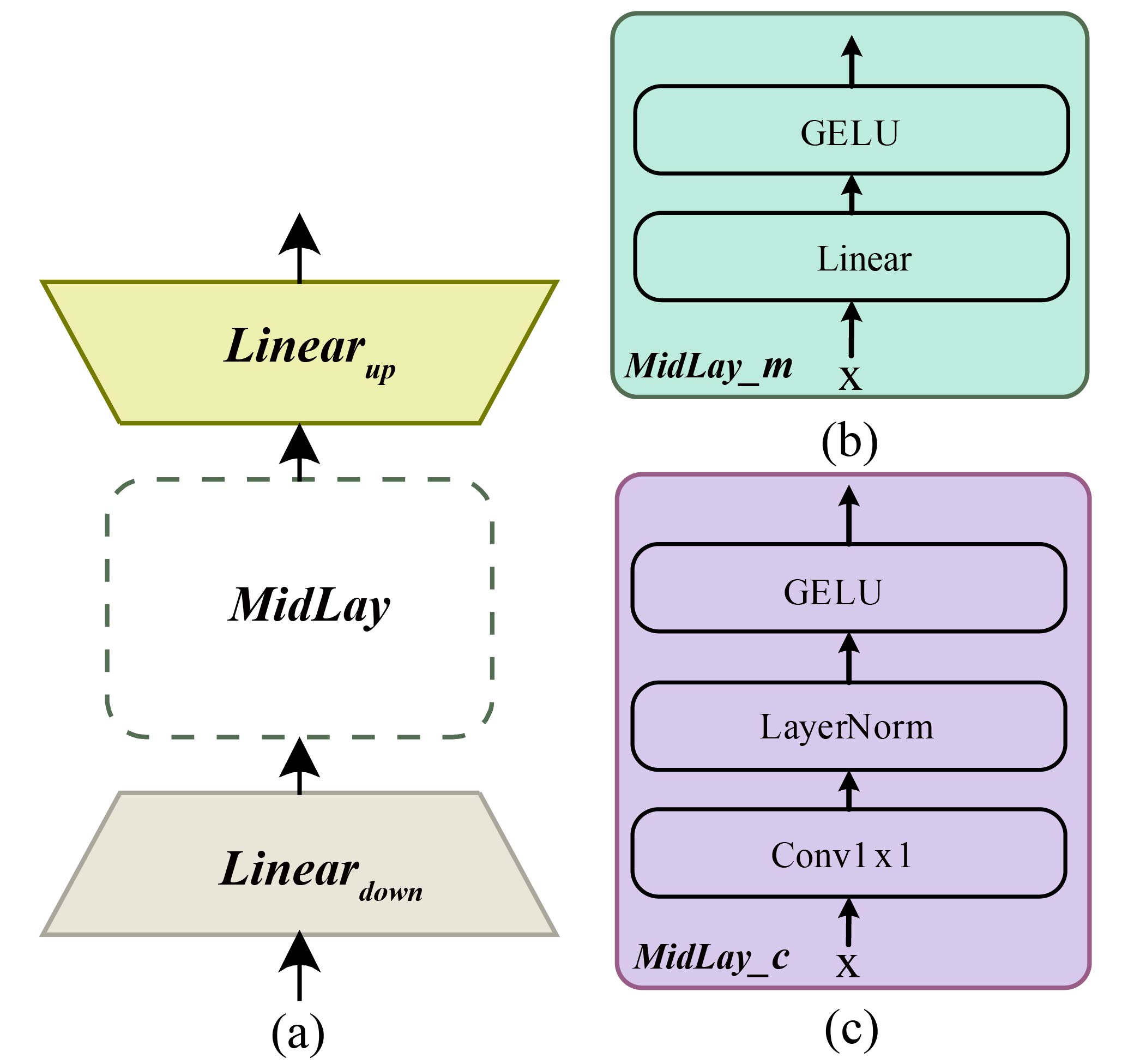}
\caption{Adaptive transfer learning structure. (a) ATL structure. (b) MidLay\_m structure. (c) MidLay\_c structure.}
\label{fig2}
\end{figure}

In addition, we designed four other ATL structures by replacing different MidLays, as shown in Fig. 3, which are 2-MidLay\_m, 2-MidLay\_c, 3-MidLay\_c,and 2-MidLay\_m+3-MidLay\_c. The ATL computation process for the structures of 2-MidLay\_m and 2-MidLay\_c, respectively, is specifically denoted as:

\begin{align} 
\label{deqn_ex5}
{\varTheta }_{i}&={\mathcal{L}}_{up}\left ( \mathcal{G}\left ( \mathcal{L}\left ( \mathcal{G}\left ( \mathcal{L}\left ( {\mathcal{L}}_{down}\left ( {\Phi }_{i}\right )\right )\right )\right )\right )\right )
\end{align}

\begin{figure}[!t]
\centering
\includegraphics[width=2.5in]{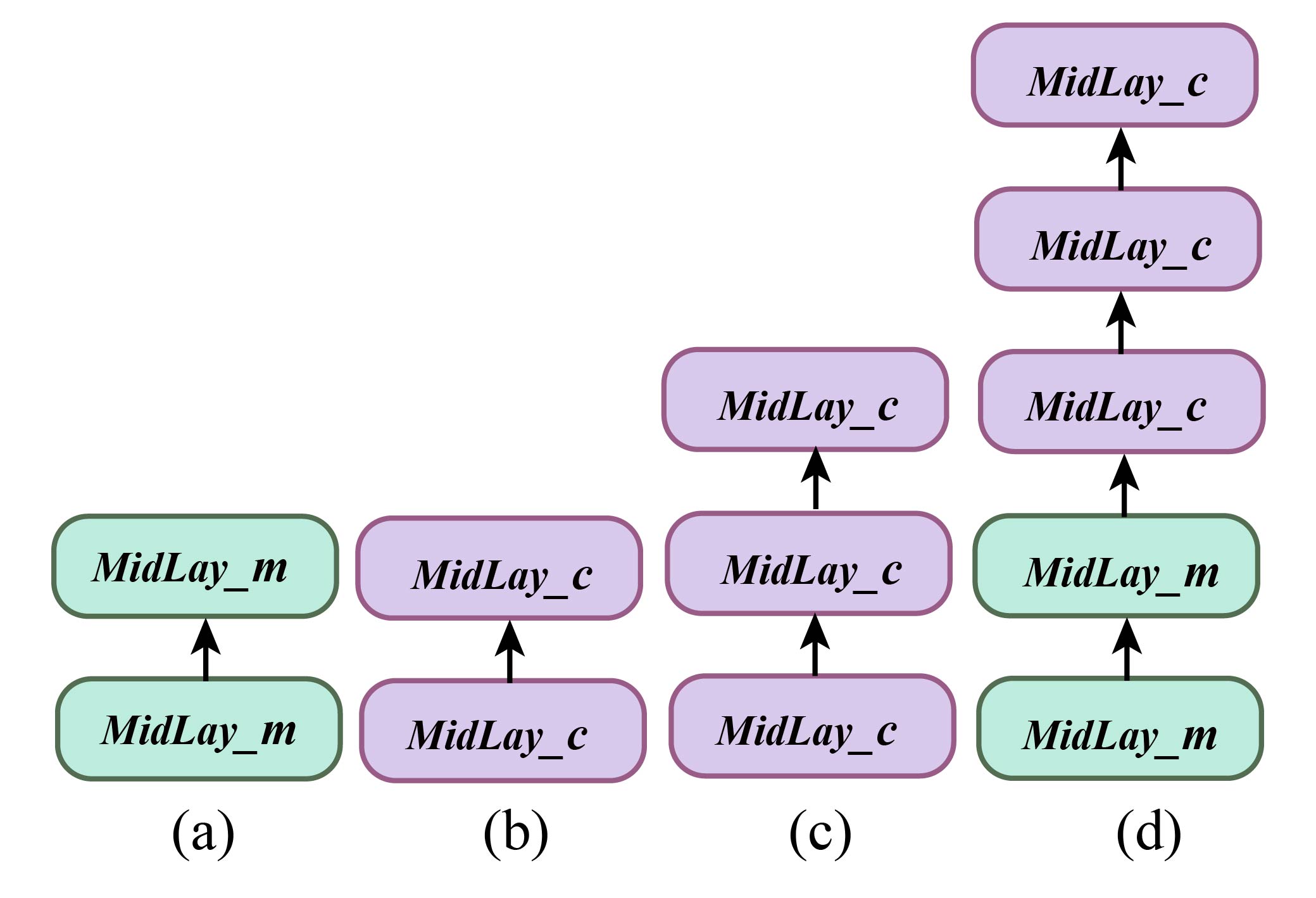}
\caption{Four different adaptive transfer learning structures. (a) 2\-MidLay\_m. (b) 2\-MidLay\_c. (c) 3\-MidLay\_c. (d) 2\-MidLay\_m + 3\-MidLay\_c.}
\label{fig3}
\end{figure}

\subsection{ATL deployment locations}
In order to study the effect of ATL deployment location on the accuracy of TransLandSeg, as in Fig. 4(b), we try to deploy ATL inside the Transformer block. Specifically, an ATL layer is inserted between the MSA and MLP layers of each Transformer block. The detailed procedure is described below: 
\begin{align} 
\label{deqn_ex6}
{x'}_{A}&={\varOmega }_{tune}\left ( {x}_{A}\right )\\
{x}_{T}&={x'}_{A}+x'
\end{align}
Where ${x}_{A}$ denotes the output of MSA, ${x'}_{A}$ denotes the output of ATL, and ${x}_{T}$ is the result after residual connection join fusion, which is subsequently sent to LayerNorm and a MLP block.
\begin{equation} 
\label{deqn_ex7} \tau = MLP\left ( LN\left ( {x}_{T}\right )\right )+{x}_{T}
\end{equation}

The training process is shown in Fig. 4(c), where only the ATL layer and mask decoder are trained and the Transformer layer and patch embedding are frozen.

\begin{figure*}[!t] 
\centering 
\includegraphics[width=7.5in]{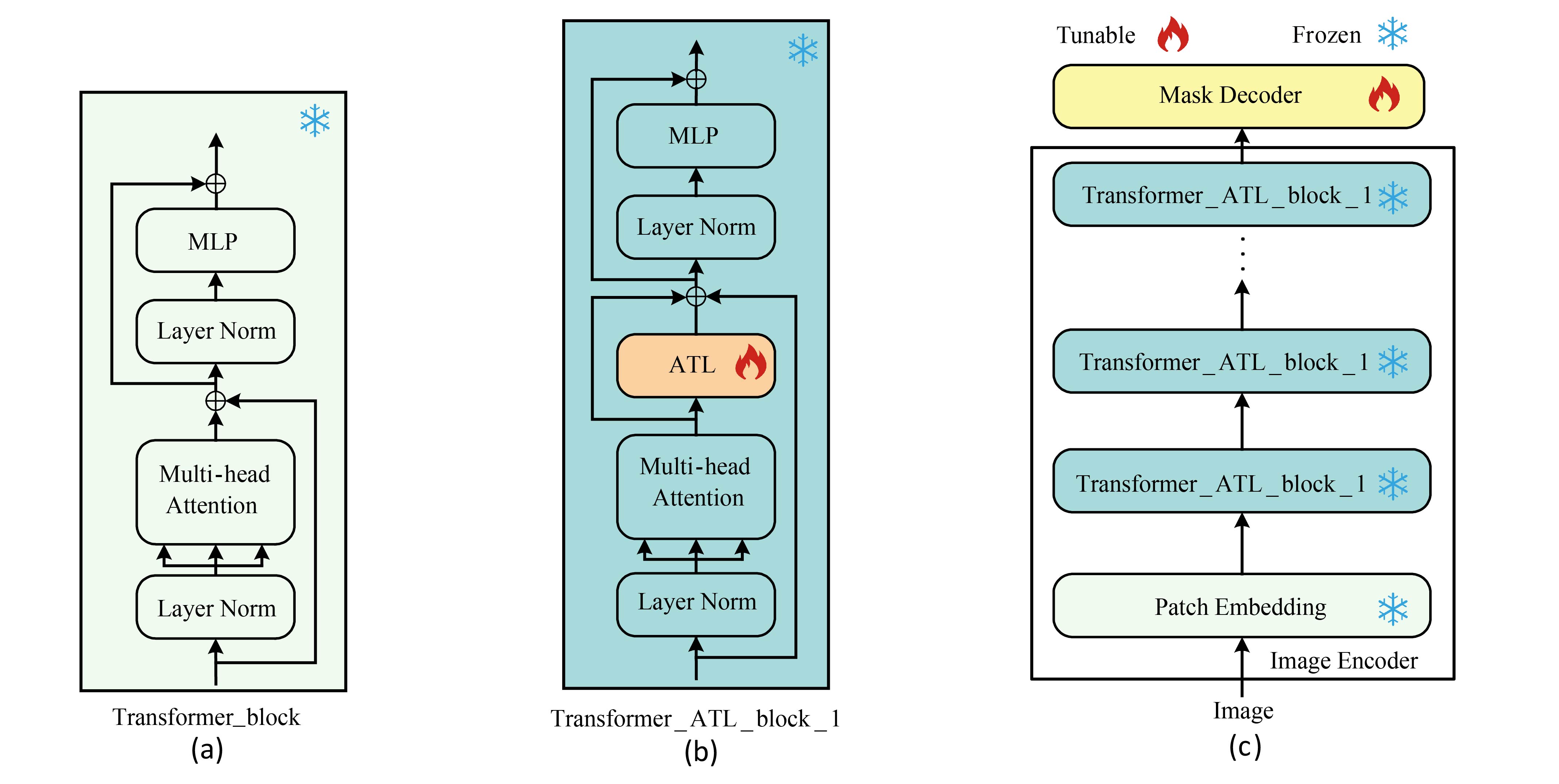} 
\caption{ATL placed inside the Transformer block. (a) The original Transformer block structure in SAM's image encoder. (b) Structure of ATL placed inside the Transformer block. (c) ATL placed inside the Transformer block to train the model.} 
\label{fig4} 
\end{figure*}

\subsection{ATL without residual structure}
In order to study the effect of residual connection on the accuracy of TransLandSeg, we remove the residual connection in the ATL, specifically deploy the ATL outside the Transformer block, and the output of the ATL layer is directly input into the Transformer block, as shown in Fig. 5(a), and the detailed process is as follows:

\begin{align} 
\label{deqn_ex9}
{\Phi }_{1}&= {\tau }_{1}\left ( {\varTheta }_{tune}\left ( x'\right )\right )
\end{align}

Deploying the ATL inside the Transformer block, the output of the MSA does not go through the residual connection and goes directly to the ATL layer for subsequent computation. The detailed process is as follows:
\begin{align} 
\label{deqn_ex10}
{x'}_{A}&={\varTheta}_{tune}\left ( {x}_{A}\right )
\end{align}
\begin{figure*}[!t] 
\centering 
\includegraphics[width=7.5in]{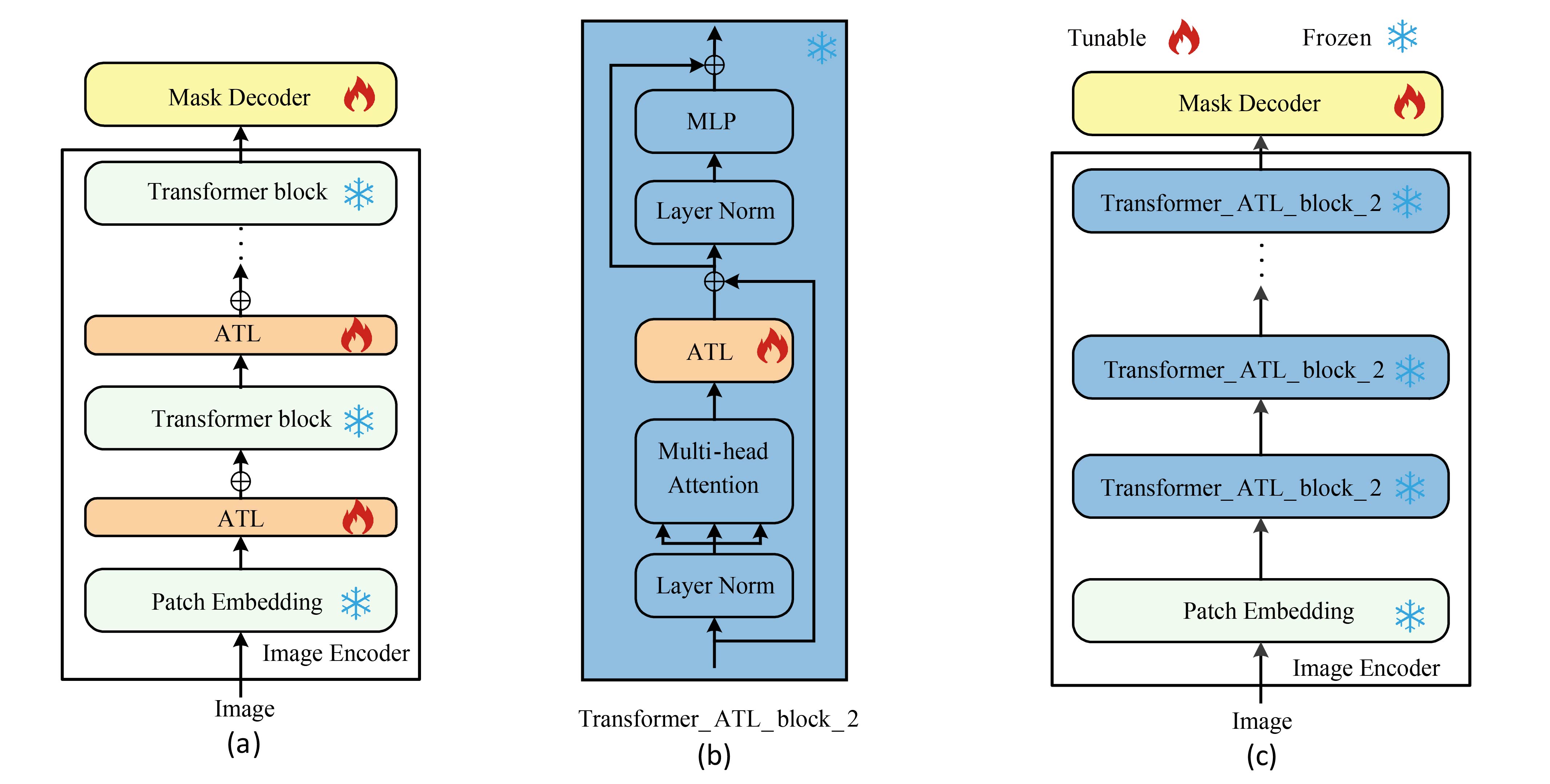} 
\caption{ATL without residual connection. (a) ATL without residual connection placed outside the Transformer block. (b) Schematic of ATL without residual connection placed inside the Transformer block. (c) ATL without residual connection placed inside the Transformer block to train the model.} 
\label{fig5} 
\end{figure*}

\section{Experimentation and Analysis}
In this section, we evaluate the proposed TransLandSeg model and compare it with several state-of-the-art reference models.
\subsection{Datasets}
We chose the Landslide4Sense dataset and the Bijie landslide dataset for our experiments. The Landslide4Sense dataset is derived from optical imagery from the Sentinel-2 satellite and contains 3799 training samples. The spatial resolution is 10 m. The dataset selects the four most landslide-prone areas globally, which are the southern part of Hokkaido, Japan; the Kodagu district of Karnataka, India; the northern part of Kathmandu, Nepal; and the western part of Taitung County, Taiwan Province, China. The Bijie landslide dataset is derived from optical imagery from the Triple Sat satellite, acquired from May to August 2018, and contains 770 landslide images within Bijie City in northwestern Guizhou Province, China. The spatial resolution is 0.8 m. We use the Landslide4Sense dataset and the Bijie landslide dataset in different semantic segmentation models for training and segmentation performance evaluation. We use the Bijie landslide dataset for training and segmentation performance evaluation in ten different TransLandSeg model structures.

According to the needs of the model structure, in our experiments, we adjusted the image size to 1024 × 1024 pixels and selected three bands of RGB for each image. In Fig. 6, several images with different morphological features and their labels were selected from each of the two datasets for display.  

\begin{figure*}[!t] 
\centering 
\includegraphics[width=7.5in]{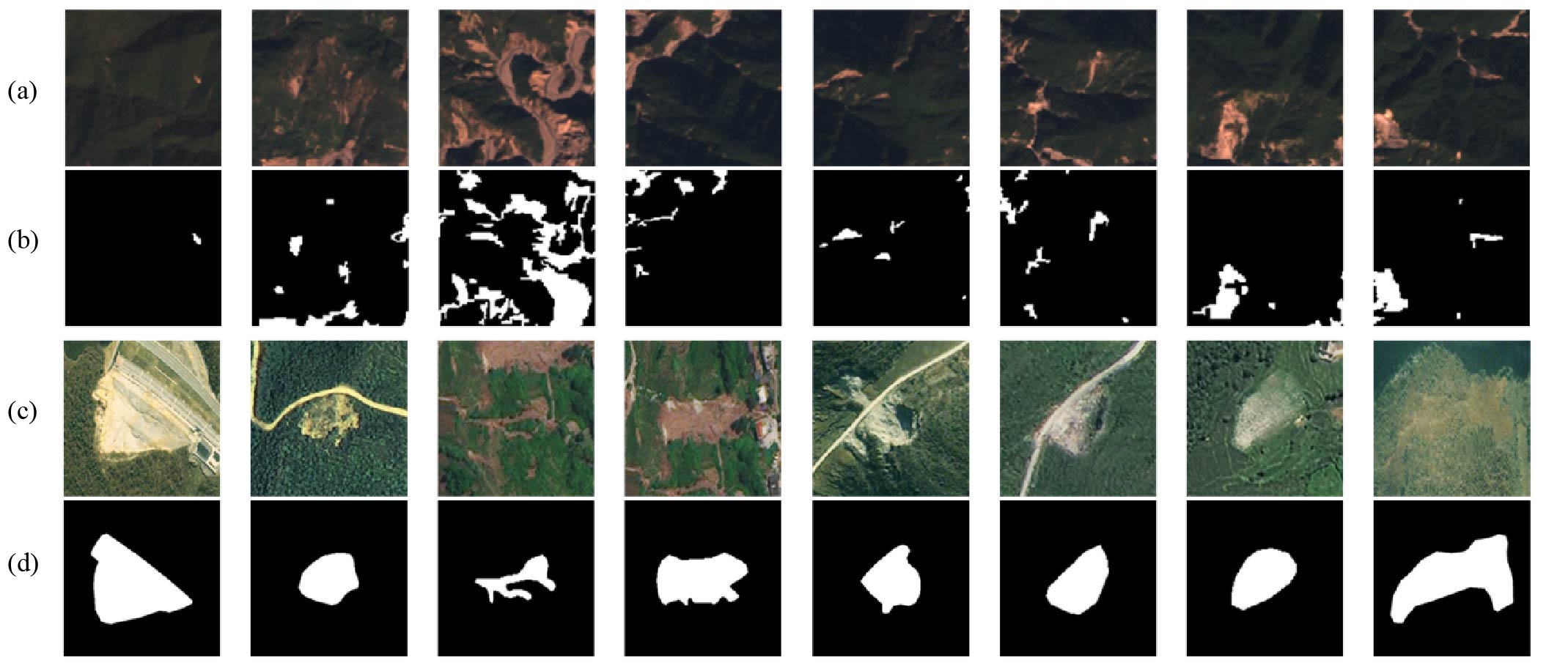} 
\caption{Visualization results of Landslide4Sense dataset and Bijie landslide dataset. (a)(b) are the Landslide4Sense dataset image and its labels. (c)(d) are the Bijie landslide dataset image and its labels.} 
\label{fig6} 
\end{figure*}

\subsection{Evaluation Metrics}
The metrics to evaluate the segmentation performance are Precision(P), Recall(REC), F1-score, Overall Accuracy(OA), Intersection over Union (IoU) and Mean Intersection over Union (MIoU). The equations are listed as follows:

\begin{equation} 
\label{deqn_ex11} P= \frac{TP}{TP+FP}
\end{equation}

\begin{equation} 
\label{deqn_ex12} REC= \frac{TP}{TP+FN}
\end{equation}

\begin{equation} 
\label{deqn_ex13} F1= \frac{ 2\times P\times REC}{P+REC}
\end{equation}

\begin{equation} 
\label{deqn_ex14} OA= \frac{TP+TN}{TP+TN+FN+FP}
\end{equation}

\begin{equation} 
\label{deqn_ex15} IoU= \frac{TP}{TP+TN+FP}
\end{equation}

\begin{equation} 
\label{deqn_ex16} MIoU=\frac{\frac{TP}{TP+FN+FP}+\frac{TN}{TN+FN+FP}}{2}
\end{equation}
where TP, FP, TN and FN are “True Positive”, “False Positive”, “True Negative” and “False Negative”, respectively.

\subsection{Implementation Details}
We use the ViT-L version of SAM. BCE loss and dice loss are used for all semantic segmentation. The AdamW optimizer is used for all the experiments. The initial learning rate is set to 2e-4. Cosine decay is applied to the learning rate. The training of all semantic segmentation is performed for 50 epochs. The experiments are implemented using PyTorch on one NVIDIA Tesla A100 GPU.
\subsection{Experimental Results}
In order to validate the overall performance of our proposed TransLandSeg model, we compared the TransLandSeg model with widely used and recognized semantic segmentation methods on the Bijie landslide dataset and the Landslide4Sense dataset. These semantic segmentation methods include PSPNet, Deeplabv3+, UNet, and CCNet. We use the evaluation metrics of chapter 5.2 to assess the segmentation performance, and the quantitative results are shown in Tables 1 and 2, respectively.
\subsubsection{Results on the Bijie landslide dataset} 
Table 1 lists the results of each semantic segmentation method. The results show that the TransLandSeg proposed in this paper outperforms the other methods in all the metrics. PSPNet uses the Pyramid Pooling Module (PPM) to pool the original feature maps and combine spatial details of different sizes to obtain different scales. The results show that PSPNet achieves 80.52\% of MIoU and 88.53\% of F1. Deeplabv3+ extends the receptive field by employing extended convolution, thus improving the model's ability to extract global contextual information. The results show that Deeplabv3+ achieves 80.28\% of MIoU and 88.29\% of F1. UNet continuously integrates spatial information of the underlying features by jumping connections, and its segmentation is slightly worse than that of Deeplab V3+, which achieves 76.2\% of MIoU and 85.29\% of F1. CCNet to get context information on the longitudinal and transversal paths through the longitudinal and transversal networks, achieving 84.43\% of F1 and 75.09\% of MIoU. In contrast, our proposed TransLandSeg model has powerful feature extraction capabilities and can achieve excellent performance. Ultimately, the TransLandSeg model obtains 88.10\% MIoU and 93.41\% F1, which improves 7.58\% MIoU and 4.88\% F1 over the second-best PSPNet model.

Figure 7 shows the prediction results of several semantic segmentation methods involved in Table 1. It can be shown that TransLandSeg achieves excellent results for both the extraction of landslide detail information and landslide detection. In the first three rows of Fig. 7, the landslide is more exposed, and all models are able to identify the landslide, but some models are missing part of the landslide information at the margins and are unable to accurately segment the landslide range, and TransLandSeg shows its advantages in the extraction of detail information. In addition, in the first and third rows of Figure 7, some models will wrongly identify other bare areas not covered by plants (e.g., roads) as part of the landslide as well, and TransLandSeg does not make such mistakes. For hard-to-distinguish features and blurred images, TransLandSeg can arrive at an accurate judgment by virtue of SAM's original powerful segmentation capability and the new knowledge learned from the ATL. Even if the landslide is partially covered by plants (fourth row of Fig. 7) or the image is obscured by fog (fifth row of Fig. 7), TransLandSeg can still identify the landslide accurately and segment the landslide extent precisely. This is a task that no other model can accomplish. Overall, TransLandSeg performs well in landslide segmentation, with the ability to integrate global information and local detail information.

\begin{table*}
\centering
\caption{Comparing the results of different models on the Bijie landslide dataset}
\label{tab1}
\resizebox{\textwidth}{!}{%
  \begin{tabular}{l|c|c|c|c|c|c}
    \hline
    \textbf{Model} & \textbf{P(\%)} & \textbf{REC(\%)} & \textbf{F1-score(\%)} & \textbf{OA(\%)} & \textbf{MIoU(\%)} & \textbf{Landslide-IoU(\%)} \\
    \hline
    CCNet & 85.24 & 83.64 & 84.43 & 94.49 & 75.09 & 56.11 \\
    Deeplabv3+ & 89.19 & 87.40 & 88.29 & 95.85 & 80.28 & 65.06 \\
    UNet & 85.81 & 84.77 & 85.29  & 94.75 & 76.20 & 58.06 \\
    PSPNet & 90.77 & 86.39 & 88.53 & 96.03 & 80.52 & 65.32 \\
    \hline
    TransLandSeg & \textbf{93.14} & \textbf{93.69} & \textbf{93.41} & \textbf{97.60} & \textbf{88.10} & \textbf{78.83} \\
    \hline
  \end{tabular}%
  }
\end{table*}

\begin{figure*}[!t] 
\centering 
\includegraphics[width=7.5in]{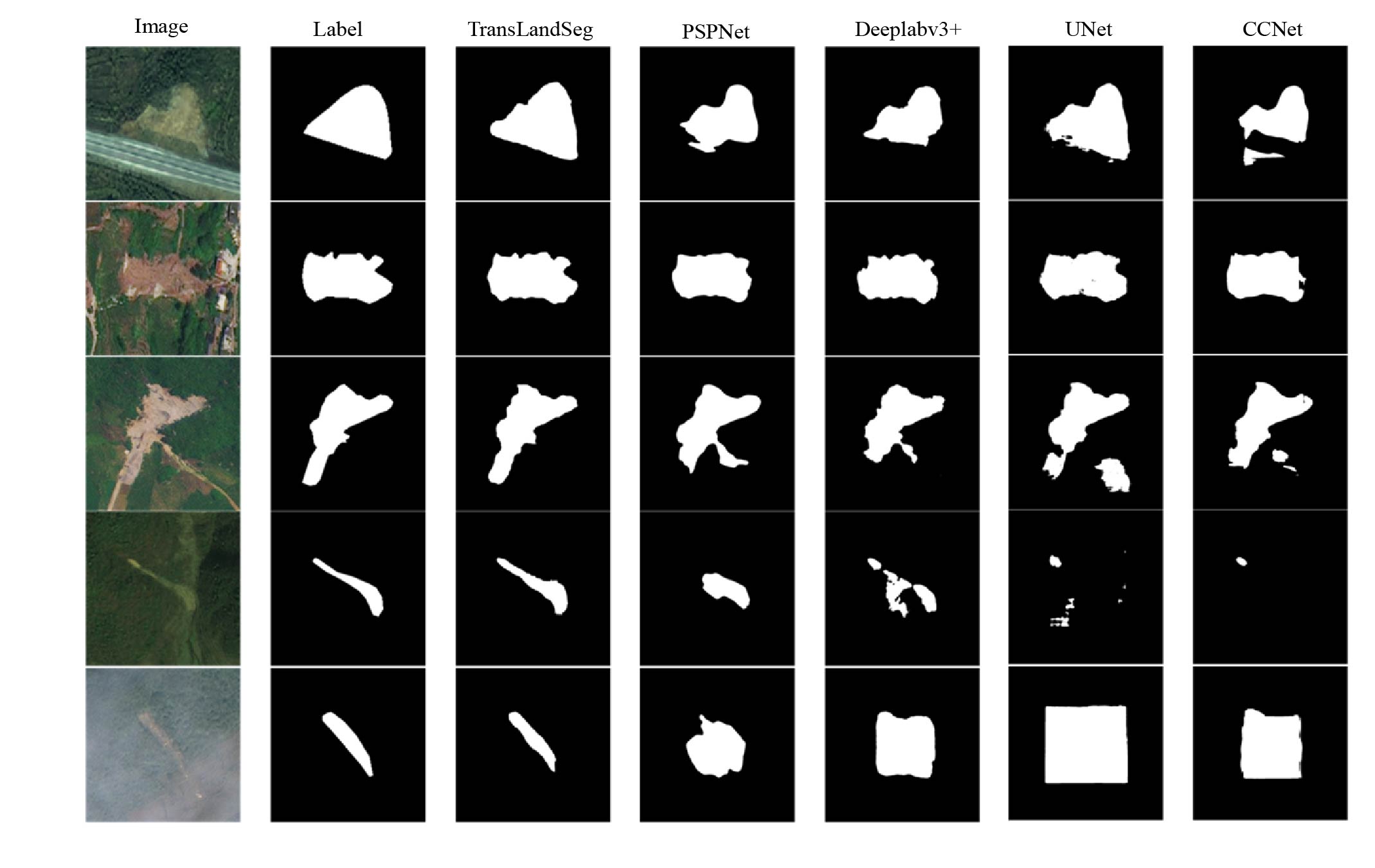} 
\caption{Comparing the segmentation results of different models on the Bijie landslide dataset.} 
\label{fig7} 
\end{figure*}

\subsubsection{Results on Landslide4Sense dataset} 
Table 2 gives the segmentation results of each method on the Landslide4Sense dataset, which further proves the effectiveness of the proposed TransLandSeg. Our TransLandSeg achieves 75.99\% on MIoU and 84.46\% on F1, which is better than other methods. The segmentation accuracy on the Bijie landslide dataset is generally higher than that on the Landslide4Sense dataset due to the different data resolutions. It is worth noting that Deeplabv3+ performs significantly worse than UNet, which verifies that due to the differences between the two networks, the UNet network performs better than Deeplabv3+ when dealing with complex and diverse small targets. Compared to other comparative methods, our TransLandSegis has the highest score in every metric. It improves MIoU by 7.78\% and F1 by 6.97\% over the second-best PSPNet model.

We visualize the segmentation results as shown in Fig. 8. Observing the first three rows, identifying densely distributed small-scale landslides is challenging, and this task is more demanding for the model in terms of local detail information extraction. The TransLandSeg model can accurately identify dense and small-scale landslide targets through its powerful information extraction capability. In the fourth to fifth rows, we can find that TransLandSeg also works well for large and continuous landslide detection, which reflects the model's ability to extract global contextual information. Meanwhile, due to the limitation of image resolution, it is difficult to recognize the target when it is more covered by vegetation, and greater ability to extract detailed information is needed.
\begin{table*}
\centering
\caption{Comparing the results of different models on the Landslide4Sense dataset.}
\label{tab2}
\resizebox{\textwidth}{!}{%
  \begin{tabular}{l|c|c|c|c|c|c}
    \hline
    \textbf{Model} & \textbf{P(\%)} & \textbf{REC(\%)} & \textbf{F1-score(\%)} & \textbf{OA(\%)} & \textbf{MIoU(\%)} & \textbf{Landslide-IoU(\%)} \\
    \hline
    CCNet & 81.55 & 70.75 & 75.77 & 98.08 & 66.11 & 34.15 \\
    Deeplabv3+ & 71.72 & 74.98 & 73.31 & 97.34 & 64.37 & 31.42 \\
    UNet & 75.22 & 78.55 & 76.85 & 97.71 & 67.67 & 37.67 \\
    PSPNet & 75.18 & 79.94 & 77.49 & 97.71 & 68.21 & 38.75 \\
    \hline
    TransLandSeg & \textbf{84.99} & \textbf{83.93} & \textbf{84.46} & \textbf{98.59} & \textbf{75.99} & \textbf{53.41} \\
    \hline
  \end{tabular}%
  }
\end{table*}

\begin{figure*}[!t] 
\centering 
\includegraphics[width=7.5in]{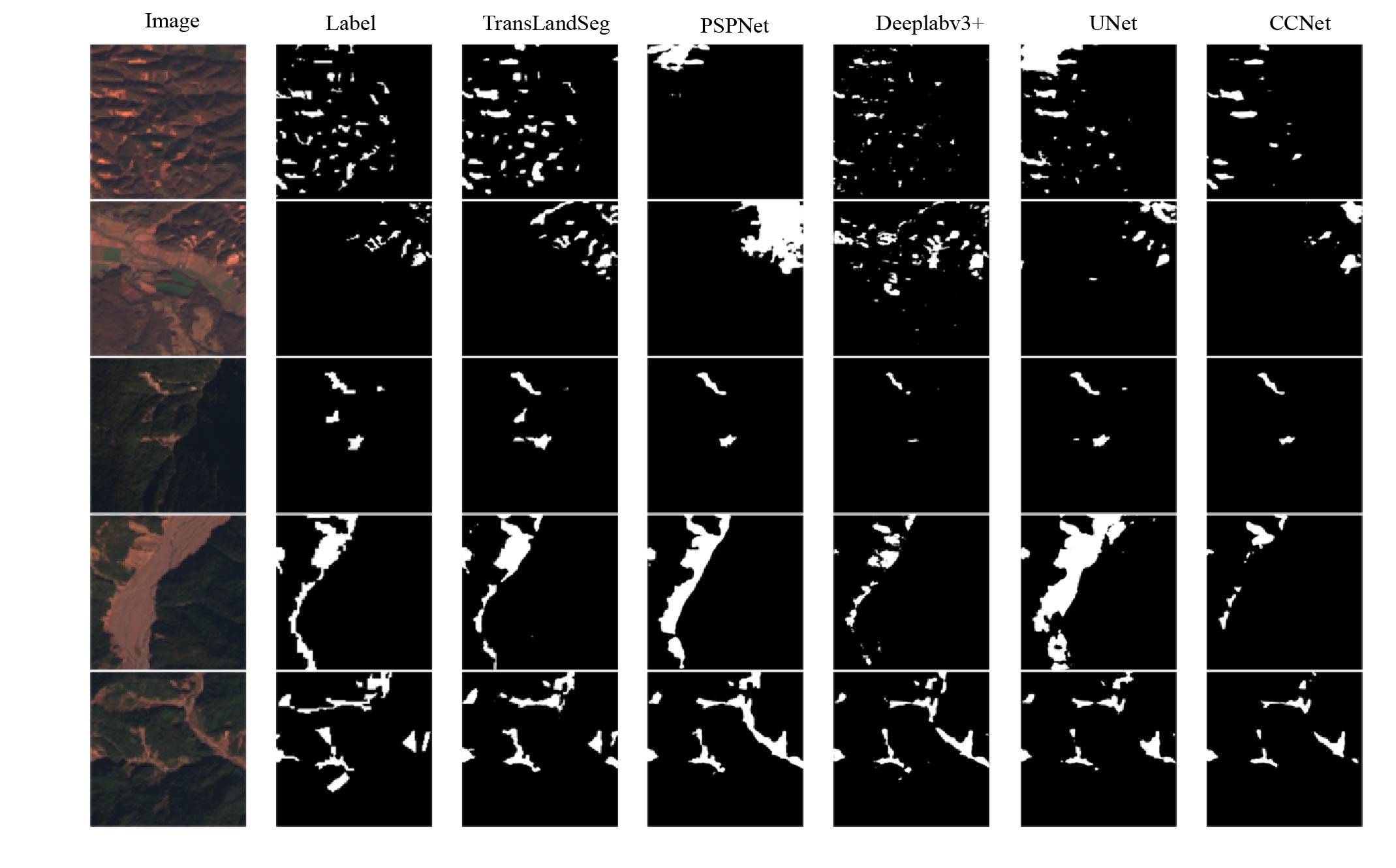} 
\caption{Comparing the segmentation results of different models on the Landslide4Sense dataset.} 
\label{fig8} 
\end{figure*}

\section{Discussion}
In this section, we discuss the superiority of the TransLandSeg model and the effect of different ATL structures on the accuracy of TransLandSeg.
\subsection{Superiority of TransLandSeg compared to other models}
TransLandSeg achieves high accuracy in landslide detection compared to other traditional semantic segmentation models due to the fact that TransLandSeg successfully transfers the powerful extraction capabilities of SAM to itself. As shown in Table 3, the Train Parameters of TransLandSeg are only 1.3\% of the Total Parameters. TransLandSeg's Train Parameters are only about 5-10\% of other models. This is because we freeze most of the parameters of the model and train only the ATL and mask decoder, which greatly saves computational resources. Comprehensively, as shown in Figure 9, TransLandSeg outperforms the other methods in both performance and efficiency. Therefore, we guess that the trend will be to use the foundation model to adapt to specific downstream tasks. On the one hand,  this approach will ensure that the model will have high accuracy, and on the other hand, it will avoid the trouble of training the model from zero. 
\begin{table}[]
\caption{Comparing the Total Parameters and Train Parameters of different models.}
\begin{tabular}{l|l|l}
\hline
\textbf{Model} & \textbf{Total Parameters/(M)} & \textbf{Train Parameters/(M)} \\ \hline
CCNet          & 71.27                         & 71.27                         \\
Deeplabv3+     & 40.35                         & 40.35                         \\
UNet           & 75.36                         & 75.36                         \\
PSPNet         & 46.58                         & 46.58                         \\ \hline
TransLandSeg   & \textbf{312.45}               & \textbf{4.18}                 \\ \hline
\end{tabular}
\end{table}

\begin{figure}[!t]
\centering
\includegraphics[width=2.5in]{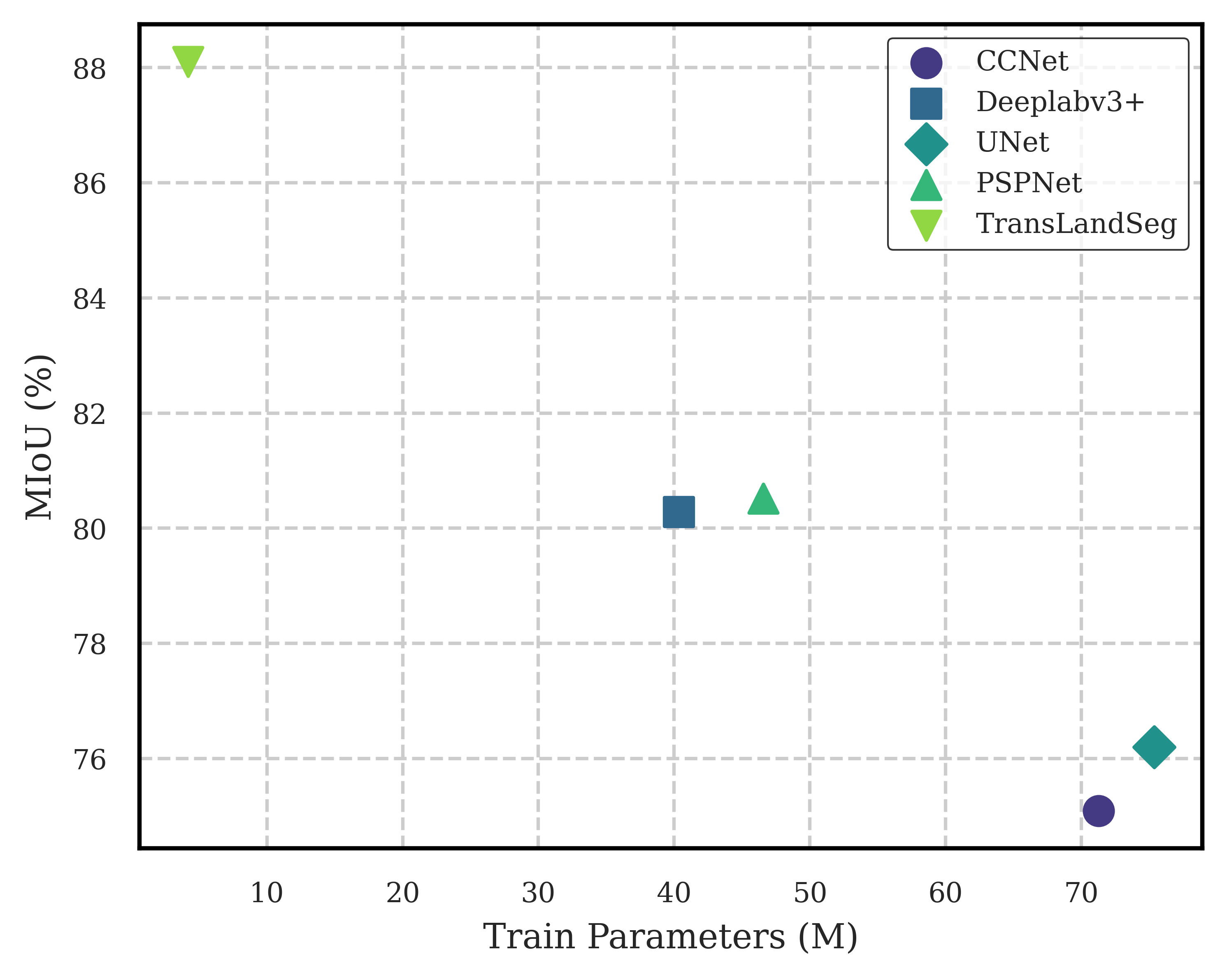}
\caption{Comparing the Train Parameters and MIoU of different models.}
\label{fig9}
\end{figure}
\subsection{Superiority of TransLandSeg compared to SAM}
SAM was trained using over 10 million images, and it demonstrates strong segmentation capabilities on natural images. We tested SAM in a number of landslide segmentation tasks, performing landslide segmentation experiments using SAM's own point prompts and auto-segmentation features. The experimental results show that SAM underperforms in this task. In the first three rows of Fig. 10, SAM is almost unable to recognize landslide features by point prompts. In the last two rows of Fig. 10, SAM can recognize landslides by point prompts but cannot perform a complete segmentation of landslides. If the automatic segmentation mode is used, it is more difficult to extract the landslide information, and the segmentation results cannot meet practical needs.

The main reason for the failure of SAM is the lack of remote sensing images in the pre-training data. In order to adapt SAM to the landslide segmentation task, we use ATL to achieve this goal. As shown in Fig. 10, with TransLandSeg, we are able to significantly improve the landslide detection effect, and TransLandSeg is able to recognize the landslide clearly and segment the landslide range accurately. This experiment shows that TransLandSeg has excellent results in landslide segmentation. We believe that the success of TransLandSeg is due to the fact that the capabilities of the original base model have been retained and infused with new knowledge. The original capabilities and the new knowledge have been perfectly integrated.

\begin{figure*}[!t] 
\centering 
\includegraphics[width=7.5in]{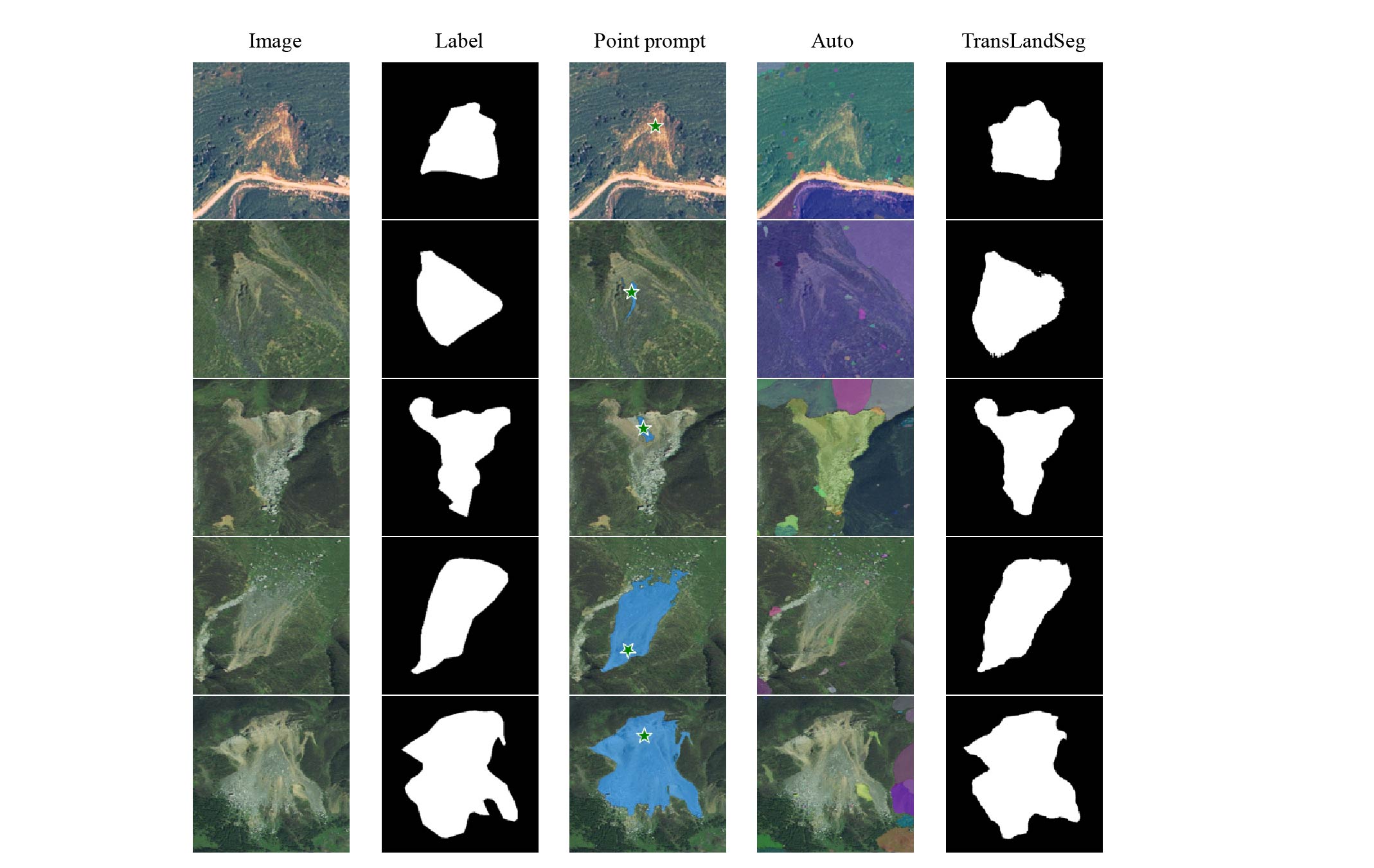} 
\caption{Visualization of landslide segmentation with different prompting  methods of SAM.} 
\label{fig10} 
\end{figure*}

\subsection{Effect of different ATL structures on the accuracy of TransLandSeg}
In order to study the effect of different ATL structures on the accuracy of TransLandSeg, we performed ablation experiments on the Bijie dataset using TransLandSeg as the baseline network. Firstly, we study the effect of five different MidLay structures on the accuracy of TransLandSeg. 
The five different MidLay structures are shown in Fig. 2, Fig. 3. Secondly we study the effect of ATL deployment location on the accuracy of TransLandSeg, the structure of which is shown in Fig. 1, Fig. 4. Finally, we study the effect of the residual connection on the accuracy of TransLandSeg, the structure is shown in Fig. 1, Fig. 5.

\subsubsection{MidLay structure}
As shown in Table 4, we compare TransLandSeg with TransLandSeg-1, TransLandSeg-2, TransLandSeg-4, and TransLandSeg-5. The results show that the MidLay structure does not have much effect on TransLandSeg, and increasing the number of MidLay layers and the complexity of the structure does not improve the detection accuracy of the model.
\subsubsection{ATL Deployment Location}
As shown in Table 4, we compare TransLandSeg with TransLandSeg-8, compare TransLandSeg-5 with TransLandSeg-6. The results show that the accuracy index of ATL deployed outside the Transformer block is higher than that placed inside the Transformer block under the same other conditions, and different ATL structures have different degrees of influence.
\subsubsection{Residual connection}
As shown in Table 4, we compare TransLandSeg-2 with TransLandSeg-3, compare TransLandSeg with TransLandSeg-7. The results show that the effect of residual connection on the fine-tuning effect is huge, and the accuracy of the model with residual connection is much higher than that without residual connection. It also shows that residual connection plays a vital role in ATL.

Summarizing the above results, our proposed TransLandSeg, i.e., ATL, has the best results using 1 layer of MidLay\_m with residual connection and placed outside the Transformer block.

\begin{table*}
\centering
\caption{Evaluation of the accuracy of landslide identification by different SAL models.}
\label{tab3}
\resizebox{\textwidth}{!}{%
\begin{tabular}{c|c|cc|cccccc}
\hline
\multirow{2}{*}{\makecell{Model}} & \multirow{2}{*}{\makecell{MidLay}}   & \multicolumn{2}{c|}{\makecell{Modules}}                            & \multirow{2}{*}{\makecell{P(\%)}} & \multirow{2}{*}{\makecell{REC(\%)}} & \multirow{2}{*}{\makecell{F1-score(\%)}} & \multirow{2}{*}{\makecell{OA(\%)}} & \multirow{2}{*}{\makecell{MIoU(\%)}} & \multirow{2}{*}{\makecell{Landslide-IoU(\%)}} \\ \cline{3-4}
                       &                               & \multicolumn{1}{c|}{\makecell{residue connection}} & \makecell{outside block} &                        &                          &                               &                         &                           &                                    \\ \hline
TransLandSeg-1                  & MidLay\_c$\times$3                  & $\checkmark$                           & $\checkmark$  & 92.26                  & 90.71                    & 91.48                         & 96.97                   & 85.01                     & 73.33                              \\ \cline{1-2}
TransLandSeg-2                  & \multirow{2}{*}{MidLay\_c$\times$2} & $\checkmark$                           & $\checkmark$  & 91.43                  & 88.95                    & 90.17                          & 96.54                   & 83.01                     & 69.77                              \\ \cline{1-1}
TransLandSeg-3                  &                               & $\times$                               & $\checkmark$  & 84.2                   & 83.87                    & 84.03                         & 94.25                   & 74.59                     & 55.36                              \\ \cline{1-2}
TransLandSeg-4                  & MidLay\_m$\times$2                   & $\checkmark$                           & $\checkmark$  & \textbf{93.91}         & 92.24                    & 93.07                         & 97.54                   & 87.52                     & 77.74                              \\ \cline{1-2}
TransLandSeg-5                  & \multirow{2}{*}{MidLay\_m$\times$2 + MidLay\_c$\times$3} & $\checkmark$          & $\checkmark$  & 92.49                  & 91.89                    & 92.19                         & 97.19                   & 86.13                     & 75.33                              \\ \cline{1-1}
TransLandSeg-6                  &                               & $\checkmark$                           & $\times$      & 88.56                  & 86.29                    & 87.41                         & 95.57                   & 79.05                     & 62.88                              \\ \cline{1-2}
TransLandSeg-7                  & \multirow{3}{*}{MidLay\_m$\times$1}  & $\times$                               & $\checkmark$  & 81.39                  & 85.06                    & 83.18                         & 93.52                   & 73.3                      & 53.6                               \\ \cline{1-1}
TransLandSeg-8                  &                               & $\checkmark$                           & $\times$      & 91.92                  & 91.15                    & 91.53                         & 96.96                   & 85.11                     & 73.53                              \\ \cline{1-1} \cline{3-10} 
TransLandSeg                    &                               & $\checkmark$                           & $\checkmark$  & 93.14                  & \textbf{93.69}           & \textbf{93.41}                & \textbf{97.6}           & \textbf{88.1}             & \textbf{78.83}                     \\ \hline
\end{tabular}%
}
\end{table*}

\section{Conclusion}
In this work, we first propose a new network for landslide detection, named TransLandSeg, which achieves better performance compared to traditional semantic segmentation methods. Second, we propose a ATL structure for VFM, which not only achieves successful transfer from the original model to the downstream task but also greatly improves the training efficiency of the model. Finally, we explored the effect of different ATL structures on the accuracy of TransLandSeg through ablation experiments, and provided principled suggestions for the design of adapter-tuning layers. We hope that our work can provide a reference for VFM applications in remote sensing.SAM can not only use ATL to adapt downstream tasks, but removing the original prompt also loses the dynamic interaction capability of the model. This thesis only solves one kind of downstream task adaptation problem. To form a generalized model for remote sensing, fine-tuning massive remote sensing data is still needed. The CLIP for adapting remote sensing data is trained to realize cross-modal remote sensing dynamic recognition.

\section*{Acknowledgments}
We extend our heartfelt thanks to Heywhale, who provided the arithmetic platform for this work.

\vfill


\begin{thebibliography}{1}
\bibliographystyle{IEEEtran}

\bibitem{ref1}
Z. Lu, Y. Peng, W. Li, J. Yu, D. Ge, and W. Xiang, “An Iterative Classification and Semantic Segmentation Network for Old Landslide Detection Using High-Resolution Remote Sensing Images.” arXiv, Apr. 24, 2023. 

\bibitem{ref2}
U. Haque et al., “The human cost of global warming: Deadly landslides and their triggers (1995–2014),” Sci. Total Environ., vol. 682, pp. 673–684, Sep. 2019, doi: 10.1016/j.scitotenv.2019.03.415. 

\bibitem{ref3}
S. Plank, A. Twele, and S. Martinis, “Landslide Mapping in Vegetated Areas Using Change Detection Based on Optical and Polarimetric SAR Data,” Remote Sens., vol. 8, no. 4, p. 307, Apr. 2016, doi: 10.3390/rs8040307.

\bibitem{ref4}
H. Zhang, “Remote Sensing Data Processing of Urban Land Using Based on Artificial Neural Network,” Sci. Program., vol. 2022, pp. 1–8, Mar. 2022, doi: 10.1155/2022/6994179.

\bibitem{ref5}
R. N. Keyport, T. Oommen, T. R. Martha, K. S. Sajinkumar, and J. S. Gierke, “A comparative analysis of pixel- and object-based detection of landslides from very high-resolution images,” Int. J. Appl. Earth Obs. Geoinformation, vol. 64, pp. 1–11, Feb. 2018, doi: 10.1016/j.jag.2017.08.015. 

\bibitem{ref6}
A. Mondini, M. Santangelo, M. Rocchetti, E. Rossetto, A. Manconi, and O. Monserrat, “Sentinel-1 SAR Amplitude Imagery for Rapid Landslide Detection,” Remote Sens., vol. 11, no. 7, p. 760, Mar. 2019, doi: 10.3390/rs11070760.

\bibitem{ref7}
X. Tang, Z. Tu, Y. Wang, M. Liu, D. Li, and X. Fan, “Automatic Detection of Coseismic Landslides Using a New Transformer Method,” Remote Sens., vol. 14, no. 12, p. 2884, Jun. 2022, doi: 10.3390/rs14122884.

\bibitem{ref8}
X. Wang et al., “Dual Path Attention Network (DPANet) for Intelligent Identification of Wenchuan Landslides,” Remote Sens., vol. 15, no. 21, p. 5213, Nov. 2023, doi: 10.3390/rs15215213.

\bibitem{ref9}
T. Ren, W. Gong, L. Gao, F. Zhao, and Z. Cheng, “An Interpretation Approach of Ascending–Descending SAR Data for Landslide Identification,” Remote Sens., vol. 14, no. 5, p. 1299, Mar. 2022, doi: 10.3390/rs14051299. 

\bibitem{ref10}
F. Chen, B. Yu, and B. Li, “A practical trial of landslide detection from single-temporal Landsat8 images using contour-based proposals and random forest: a case study of national Nepal,” Landslides, vol. 15, no. 3, pp. 453–464, Mar. 2018, doi: 10.1007/s10346-017-0884-x.

\bibitem{ref11}
X. Liu et al., “Feature-Fusion Segmentation Network for Landslide Detection Using High-Resolution Remote Sensing Images and Digital Elevation Model Data,” IEEE Trans. Geosci. Remote Sens., vol. 61, pp. 1–14, 2023, doi: 10.1109/TGRS.2022.3233637.

\bibitem{ref12}
A. Mohan, A. K. Singh, B. Kumar, and R. Dwivedi, “Review on remote sensing methods for landslide detection using machine and deep learning,” Trans. Emerg. Telecommun. Technol., vol. 32, no. 7, p. e3998, Jul. 2021, doi: 10.1002/ett.3998. 

\bibitem{ref13}
W. Zhao, A. Li, X. Nan, Z. Zhang, and G. Lei, “Postearthquake Landslides Mapping From Landsat-8 Data for the 2015 Nepal Earthquake Using a Pixel-Based Change Detection Method,” IEEE J. Sel. Top. Appl. Earth Obs. Remote Sens., vol. 10, no. 5, pp. 1758–1768, May 2017, doi: 10.1109/JSTARS.2017.2661802.

\bibitem{ref14}
Z. Li, W. Shi, P. Lu, L. Yan, Q. Wang, and Z. Miao, “Landslide mapping from aerial photographs using change detection-based Markov random field,” Remote Sens. Environ., vol. 187, pp. 76–90, Dec. 2016, doi: 10.1016/j.rse.2016.10.008.

\bibitem{ref15}
T. Chen, J. Trinder, and R. Niu, “Object-Oriented Landslide Mapping Using ZY-3 Satellite Imagery, Random Forest and Mathematical Morphology, for the Three-Gorges Reservoir, China,” Remote Sens., vol. 9, no. 4, p. 333, Mar. 2017, doi: 10.3390/rs9040333.
\bibitem{ref16}
T.-A. Bui, P.-J. Lee, K.-Y. Lum, C. Loh, and K. Tan, “Deep Learning for Landslide Recognition in Satellite Architecture,” IEEE Access, vol. 8, pp. 143665–143678, 2020, doi: 10.1109/ACCESS.2020.3014305.
\bibitem{ref17}
E. Shelhamer, J. Long, and T. Darrell, “Fully Convolutional Networks for Semantic Segmentation,” IEEE Trans. Pattern Anal. Mach. Intell., vol. 39, no. 4, pp. 640–651, Apr. 2017, doi: 10.1109/TPAMI.2016.2572683.
\bibitem{ref18}
L. P. Soares, H. C. Dias, and C. H. Grohmann, “Landslide Segmentation with U-Net: Evaluating Different Sampling Methods and Patch Sizes.” arXiv, Jul. 13, 2020. 
\bibitem{ref19}
X. Zheng, L. Han, G. He, N. Wang, G. Wang, and L. Feng, “Semantic Segmentation Model for Wide-Area Coseismic Landslide Extraction Based on Embedded Multichannel Spectral–Topographic Feature Fusion: A Case Study of the Jiuzhaigou Ms7.0 Earthquake in Sichuan, China,” Remote Sens., vol. 15, no. 4, p. 1084, Feb. 2023, doi: 10.3390/rs15041084.
\bibitem{ref20}
Y. Ju et al., “Loess Landslide Detection Using Object Detection Algorithms in Northwest China,” Remote Sens., vol. 14, no. 5, p. 1182, Feb. 2022, doi: 10.3390/rs14051182. 
\bibitem{ref21}
B. Li and J. Li, “Methods for landslide detection based on lightweight YOLOv4 convolutional neural network,” Earth Sci. Inform., vol. 15, no. 2, pp. 765–775, Jun. 2022, doi: 10.1007/s12145-022-00764-0.
\bibitem{ref22}
H. Yu, Y. Ma, L. Wang, Y. Zhai, and X. Wang, “A landslide intelligent detection method based on CNN and RSG\_R,” in 2017 IEEE International Conference on Mechatronics and Automation (ICMA), Takamatsu, Japan: IEEE, Aug. 2017, pp. 40–44. doi: 10.1109/ICMA.2017.8015785.
\bibitem{ref23}
B. Fang, G. Chen, L. Pan, R. Kou, and L. Wang, “GAN-Based Siamese Framework for Landslide Inventory Mapping Using Bi-Temporal Optical Remote Sensing Images,” IEEE Geosci. Remote Sens. Lett., vol. 18, no. 3, pp. 391–395, Mar. 2021, doi: 10.1109/LGRS.2020.2979693.
\bibitem{ref24}
S. Ji, D. Yu, C. Shen, W. Li, and Q. Xu, “Landslide detection from an open satellite imagery and digital elevation model dataset using attention boosted convolutional neural networks,” Landslides, vol. 17, no. 6, pp. 1337–1352, Jun. 2020, doi: 10.1007/s10346-020-01353-2.
\bibitem{ref25}
S. Yang et al., “Automatic Identification of Landslides Based on Deep Learning,” Appl. Sci., vol. 12, no. 16, p. 8153, Aug. 2022, doi: 10.3390/app12168153.
\bibitem{ref26}
H. Cai, T. Chen, R. Niu, and A. Plaza, “Landslide Detection Using Densely Connected Convolutional Networks and Environmental Conditions,” IEEE J. Sel. Top. Appl. Earth Obs. Remote Sens., vol. 14, pp. 5235–5247, 2021, doi: 10.1109/JSTARS.2021.3079196.
\bibitem{ref27}
Y. Liang, Y. Zhang, Y. Li, and J. Xiong, “Automatic Identification for the Boundaries of InSAR Anomalous Deformation Areas Based on Semantic Segmentation Model,” Remote Sens., vol. 15, no. 21, p. 5262, Nov. 2023, doi: 10.3390/rs15215262.
\bibitem{ref28}
H. Zhou et al., “Time-Series InSAR with Deep-Learning-Based Topography-Dependent Atmospheric Delay Correction for Potential Landslide Detection,” Remote Sens., vol. 15, no. 22, p. 5287, Nov. 2023, doi: 10.3390/rs15225287.
\bibitem{ref29}
P. Lv, L. Ma, Q. Li, and F. Du, “ShapeFormer: A Shape-Enhanced Vision Transformer Model for Optical Remote Sensing Image Landslide Detection,” IEEE J. Sel. Top. Appl. Earth Obs. Remote Sens., vol. 16, pp. 2681–2689, 2023, doi: 10.1109/JSTARS.2023.3253769.
\bibitem{ref30}
Z. Yang, C. Xu, and L. Li, “Landslide Detection Based on ResU-Net with Transformer and CBAM Embedded: Two Examples with Geologically Different Environments,” Remote Sens., vol. 14, no. 12, p. 2885, Jun. 2022, doi: 10.3390/rs14122885.
\bibitem{ref31}
“An Image is Worth 16x16 Words: Transformers for Image Recognition at Scale,arXiv - CS - Machine Learning .” 
\bibitem{ref32}
A. Vaswani et al., “Attention Is All You Need,” arXiv:1706.03762, Aug. 2023.
\bibitem{ref33}
H. Gong et al., “Swin-Transformer-Enabled YOLOv5 with Attention Mechanism for Small Object Detection on Satellite Images,” Remote Sens., vol. 14, no. 12, p. 2861, Jun. 2022, doi: 10.3390/rs14122861.
\bibitem{ref34}
“How to train your ViT? Data, Augmentation, and Regularization in Vision Transformers,arXiv - CS - Computer Vision and Pattern Recognition.”  
\bibitem{ref35}
J. Devlin, M.-W. Chang, K. Lee, and K. Toutanova, “BERT: Pre-training of Deep Bidirectional Transformers for Language Understanding,” arXiv:1810.04805, May 2019.
\bibitem{ref36}
K. Bi, L. Xie, H. Zhang, X. Chen, X. Gu, and Q. Tian, “Pangu-Weather: A 3D High-Resolution Model for Fast and Accurate Global Weather Forecast.” arXiv, Nov. 03, 2022. 
\bibitem{ref37}
A. Kirillov et al., “Segment Anything.” arXiv, Apr. 05, 2023. 
\bibitem{ref38}
Z. Qiu, Y. Hu, H. Li, and J. Liu, “Learnable Ophthalmology SAM.” arXiv, Apr. 26, 2023. 
\bibitem{ref39}
K. Zhang and D. Liu, “Customized Segment Anything Model for Medical Image Segmentation.” arXiv, Oct. 17, 2023.  
\bibitem{ref40}
K. Chen et al., “RSPrompter: Learning to Prompt for Remote Sensing Instance Segmentation based on Visual Foundation Model.” arXiv, Nov. 29, 2023.
\bibitem{ref41}
L. P. Osco et al., “The Segment Anything Model (SAM) for Remote Sensing Applications: From Zero to One Shot.” arXiv, Oct. 31, 2023. 
\bibitem{ref42}
N. Houlsby et al., “Parameter-Efficient Transfer Learning for NLP.” arXiv, Jun. 13, 2019. 
\bibitem{ref43}
E. B. Zaken, S. Ravfogel, and Y. Goldberg, “BitFit: Simple Parameter-efficient Fine-tuning for Transformer-based Masked Language-models.” arXiv, Sep. 05, 2022.  
\bibitem{ref44}
X. He, C. Li, P. Zhang, J. Yang, and X. E. Wang, “Parameter-efficient Model Adaptation for Vision Transformers.” arXiv, Jul. 13, 2023. 
\bibitem{ref45}
S. Chen et al., “AdaptFormer: Adapting Vision Transformers for Scalable Visual Recognition.” arXiv, Oct. 14, 2022. 
\bibitem{ref46}
O. Ghorbanzadeh, H. Shahabi, A. Crivellari, S. Homayouni, T. Blaschke, and P. Ghamisi, “Landslide detection using deep learning and object-based image analysis,” Landslides, vol. 19, no. 4, pp. 929–939, Apr. 2022, doi: 10.1007/s10346-021-01843-x.
\bibitem{ref47}
H. He, C. Li, R. Yang, H. Zeng, L. Li, and Y. Zhu, “Multisource Data Fusion and Adversarial Nets for Landslide Extraction from UAV-Photogrammetry-Derived Data,” Remote Sens., vol. 14, no. 13, p. 3059, Jun. 2022, doi: 10.3390/rs14133059.
\bibitem{ref48}
R. Fu et al., “Fast Seismic Landslide Detection Based on Improved Mask R-CNN,” Remote Sens., vol. 14, no. 16, p. 3928, Aug. 2022, doi: 10.3390/rs14163928.
\bibitem{ref49}
Z. Peng, Z. Xu, Z. Zeng, X. Yang, and W. Shen, “SAM-PARSER: Fine-tuning SAM Efficiently by Parameter Space Reconstruction.” arXiv, Dec. 18, 2023. 
\bibitem{ref50}
J. Wu et al., “Medical SAM Adapter: Adapting Segment Anything Model for Medical Image Segmentation.” arXiv, Dec. 28, 2023. 
\bibitem{ref51}
Y. Li, M. Hu, and X. Yang, “Polyp-SAM: Transfer SAM for Polyp Segmentation”.
\bibitem{ref52}
“Visual Prompting: Modifying Pixel Space to Adapt Pre-trained Models,arXiv - CS - Computer Vision and Pattern Recognition .” 
\bibitem{ref53}
“Visual Prompt Tuning,arXiv - CS - Computer Vision and Pattern Recognition.”
\bibitem{ref54}
T. Chen et al., “SAM Fails to Segment Anything? -- SAM-Adapter: Adapting SAM in Underperformed Scenes: Camouflage, Shadow, Medical Image Segmentation, and More.” arXiv, May 02, 2023. 


\end{thebibliography}
\end{document}